\documentclass[preprint,review,12pt]{elsarticle}

\usepackage{amssymb}
\usepackage{amsmath}
\usepackage{hyperref}

\begin{document}

\begin{frontmatter}

%% Title, authors and addresses

\title{NoiseTrans: Point Cloud Denoising with Transformers}

\author[1]{Guangzhe Hou}

\author[1]{Guihe Qin\corref{cor1}}
\cortext[cor1]{Corresponding Author}

\author[1]{Minghui Sun}
\author[1]{Yanhua Liang}
\author[1]{Jie Yan}
\author[1]{Zhonghan Zhang}

\address[1]{Jilin University,
             Changchun,
             Jilin,
             China}

\begin{abstract}
\textit{Introduction}: Point clouds obtained from capture devices or 3D reconstruction techniques are often noisy and interfere with downstream tasks. \\
\textit{Objectives}: The paper aims to recover the underlying surface of noisy point clouds. \\
\textit{Methods}: We design a novel model, NoiseTrans, which uses transformer encoder architecture for point cloud denoising. Specifically, we obtain structural similarity of point-based point clouds with the assistance of the transformer's core self-attention mechanism. By expressing the noisy point cloud as a set of unordered vectors, we convert point clouds into point embeddings and employ Transformer to generate clean point clouds. To make the Transformer preserve details when sensing the point cloud, we design the Local Point Attention to prevent the point cloud from being over-smooth. In addition, we also propose sparse encoding, which enables the Transformer to better perceive the structural relationships of the point cloud and improve the denoising performance.\\
\textit{Results}: Experiments show that our model outperforms state-of-the-art methods in various datasets and noise environments.
\end{abstract}


\end{frontmatter}

\section{Introduction}
\label{sec1}
With the increasing availability of 3D scanning equipment and the development of 3D reconstruction techniques, point clouds are becoming more readily available. In industry, point clouds are also widely used in areas such as autonomous driving and robotics \cite{1}. However, point clouds can be corrupted by noise during acquisition due to limitations of the equipment and blurred matching of reconstruction techniques. This noise can lead to instability of the underlying structure and severely affect the downstream understanding task. At the same time, the denoising task also faces many serious challenges due to the disorderly and irregular nature of point clouds \cite{2}. Therefore, the task of point cloud denoising is crucial to point cloud processing. A well-designed denoising method should restore the basic structure of the point cloud while retaining as much detail as possible.

Point cloud denoising methods can be divided into two different categories: non-deep learning and deep learning. Non-deep learning methods have developed over decades producing a number of methods \cite{3,4,5,6,7,8,9,10,11} for different models. Depending on the implementation method, they are generally classified as local surface fitting, non-local means, sparse coding, and graph-based methods. These methods provide a variety of ideas for point cloud denoising that are of profound value. Nevertheless, non-deep learning methods are still plagued by problems such as prior knowledge and hyper-parameter settings.

In recent years, a number of point-based deep learning denoising methods \cite{12,13,14,15,16} have also been proposed with remarkable results. In earlier work \cite{12,13,14}, deep learning methods typically used graph convolution or multi-layer perceptrons to extract features from point clouds to predict the true locations of noisy points. More recently, distinctive methods such as DMR \cite{15} and Score \cite{16} have been proposed. They constructed the point cloud as a mathematical model to remove noise. The emergence of these methods has also greatly enriched problem-solving possibilities.

With the success of DETR \cite{17} on object detection tasks, Transformer has been rapidly developed in the field of computer vision. The transformer-based models have achieved outstanding results on various image tasks. Particularly in the low-level task, a range of models such as IPT \cite{18} and Uformer \cite{19} have achieved notable performance. In the 3D vision domain, PT \cite{20} creatively used Transformer for the classification and semantic segmentation of point-based point clouds. Inspired by this success, we incorporate transformer into point-based denoising tasks and achieve state-of-the-art results. To the best of our knowledge, we are the first to use the transformer to solve point-based point cloud denoising tasks.

We propose the NoiseTrans framework, the main idea of which is to use a self-attention mechanism to capture the global semantic relevance of structural information, and the permutation invariance of Transformer allows us to disregard the order of points. To enable the transformer structure to obtain structural organization and to improve denoising performance, we designed sparse encoding. Further, considering the issue of surface detail preservation, we added the Local Point Attention module. Extensive experiments demonstrate that NoiseTrans with global semantic relevance achieves state-of-the-art results with different datasets and different noise types. The main contributions of this paper are summarized as follows:

\begin{enumerate}
	\item  We propose a novel transformer-based point cloud denoising framework, named NoiseTrans, which is exactly suitable for point-based point cloud data with noise in various scenarios.
	\item We propose the point embedding module with local point attention, which allows multi-scale extraction of local features of the point cloud and emphasizes edge weights to preserve detail.
	\item We propose learnable Sparse Encoding which is permutation invariant and it can provide more information on structural relationships for the denoising task.
	\item We compare our method on raw and synthetic point cloud datasets which are added with different types of noise to existing methods and achieve state-of-the-art performance.
\end{enumerate}

This paper is organised as follows. Section~3 describes our architecture in detail. Section~4 gives the results of various different experiments. Section~5 gives a brief summary of the paper.
\section{Related work}\label{sec2}
\subsection{Non-deep learning methods}\label{subsec1}
There has been a long history of research on point cloud denoising methods. Point cloud denoising has become particularly important in recent years as technologies such as remote sensing, robotics, and autonomous driving continue to develop. Non-deep learning point cloud denoising methods can generally be divided into four categories: local-surface-fitting based methods \cite{4,5,21}, non-local means based methods \cite{22,23}, sparsity-based methods \cite{6,7,24} and graph-based methods \cite{9,10,11}. 

The most representative local surface-based adaptation method is MLS \cite{5}, which assumes that the surface of the point cloud is smooth and fits point to plane for denoising. In addition to that, other surface fitting methods have been proposed such as jet-fitting with re-projection \cite{4} and various forms of bilateral filters \cite{21} have achieved remarkable results. However, this can be over-smooth or over-sharpened at high noise levels.

Non-local means-based methods are developed from the 2D image denoising, such as \cite{22,23}, which detect the similar structure of non-local point cloud, and then remove the noise through filtering to get the point cloud without noise. These methods can retain more surface features and avoid excessive smoothing. However, a large number of parameters need to be set, which is difficult to apply to different scenarios.

Sparsity-based methods \cite{6,7,24} are generally to first solve the optimization problem of sparsity constraints to reconstruct the normal and then update the coordinate position of the point according to the reconstructed normal to denoise. MRPCA \cite{7} is a representative of this, but performance degrades at high noise levels.

The graph-based methods \cite{9,10,11} are to define the point cloud as a graph, use the points in the point cloud as nodes in the graph, and then denoise it through graph filters \cite{10,11}. In particular, GLR \cite{10} uses the Laplace regularisation of the graph as a filter for denoising and achieves high effectiveness. Under high noise levels, however, it leads to instability in the structure of the graph, resulting in a degradation of the denoising effect.
\subsection{Point-based deep learning methods}\label{subsec2}
In recent years, most of the significant advances in visual recognition have been achieved by deep learning models, especially deep Convolution Neural Networks (CNNs).However, directly processing point clouds using off-the-shelf image-based methods is not straightforward since point clouds are essentially irregular and unordered, which are not suitable to be processed directly using convolution neural networks designed for gridded features. With the deepening of 3D object classification, especially the proposal of PointNet \cite{2} and PointNet++ \cite{25}, the point-based deep learning methods have become possible, and the point cloud denoising method based on the deep learning methods has also gained more extensive attention.

PointCleanNet \cite{12}, which employs PCPNet \cite{26} (a variant of PointNet \cite{2}) as the backbone, predicted the displacement vector from noise point to object surface according to the local feature of each point. Because of the multiple iterations in the denoising process, shrinkage of the point cloud occurs. TotalDn \cite{13} proposed the unsupervised point cloud denoising for the first time. They introduced the prior knowledge that the denser the point cloud is, the closer it is to the real surface. They propose a new loss function that uses neighbors to determine the location of ground truth points without considering the noise point themselves. This method works well for flatter surfaces with lower noise levels. Due to the similarity between point cloud and graph in 3D space, graph convolution network \cite{14} is used for denoising, and positive results have also been achieved. Recently, DMR \cite{15} has shifted its attention to the physical characteristics of point clouds and reconstructed the bottom manifold of point clouds through a down-up sampling network structure to the denoise point cloud. Meanwhile, Score \cite{16} regarded the point cloud with noisy points as the distribution after convolution of the underlying manifold and noise distribution and proposed that the likelihood function of the distribution should update the position of each point by iteration of gradient rise to achieve the effect of denoising. But this approach leads to the creation of some outliers.

\subsection{Transformer methods}\label{subsec3}
Transformer \cite{27} was originally proposed as a sequence-to-sequence model for machine translation \cite{28} and consists of an encoder and a decoder. Each encoder block consists mainly of a multi-head self-attention module and a feed-forward network(FFN). Compared to the encoder blocks, decoder blocks additionally insert cross-attention modules between the multi-head self-attention modules and the FFN \cite{29}. With the continuous development of natural language processing(NLP) techniques, transformer-based pre-training models have been proposed one after another. Among them, pre-trained models such as the BERT \cite{30} and GPT series \cite{31,32} have achieved advanced performance in various tasks. As a result, the transformer has become the architecture of choice for NLP.

Inspired by the great success of the self-attention mechanism in NLP, Transformer has also been transplanted to the computer vision field \cite{18,33,34}. ViT \cite{35} proposes an image transformer structure that takes image patches as input and achieves advanced results on image classification tasks. In addition, it achieves impressive performance on tasks such as object detection \cite{17,36}, semantic segmentation \cite{37} and target tracking \cite{38}. Not only that, Transformer also provides new thinking for point-based 3D point cloud processing, e.g. \cite{20} and \cite{39}, etc.

Unlike the denoising methods described above, our NoiseTrans is designed on the basis of the transformer encoder architecture. We incorporate the core self-attention mechanism into the point-based point cloud denoising method in order to achieve optimal performance.

\section{Approach}  \label{sec3}
In this section, we propose a network architecture for point cloud denoising based on the characteristics of point cloud noise and present our design considerations.
\subsection{Overview} \label{overview}
Currently, deep learning denoising methods generally fit clean surfaces either through local features \cite{12,13,14} or abstracting point clouds into mathematical models \cite{15,16}. In contrast, we focus on the global relationship of the point cloud. Previous work \cite{22} has shown that synthetic and raw point clouds tend to exhibit self-similarity. Extracting and exploiting this similar geometry and overall structure can provide a powerful means of denoising. Meanwhile, Transformer has proven through extensive experiments that its self-attention mechanism has the ability to capture long-range dependencies between elements, whether in the field of natural language processing \cite{27,30,32} or computer vision \cite{20,33,35}. This is exactly what we need for denoising tasks. Moreover, its permutation invariance is naturally suitable for disordered point clouds. Further, the unique point-to-point correspondence of the transformer structure is convenient and crucial for our denoising task.

\begin{figure}[h]
	\centering{\includegraphics[scale=0.23]{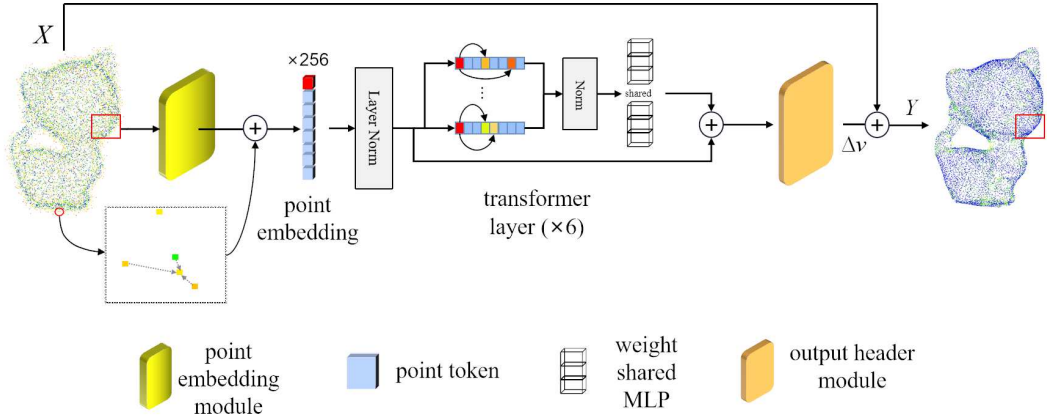}}
	\caption{Illustration of the proposed point cloud denoising framework. The different colors of the token show the degree of effect on the red token.}
	\label{fig1}
	\vskip-5pt
\end{figure}

Based on the above analysis, we propose a transformer-based point cloud denoising method, NoiseTrans, which is shown in Figure~\ref{fig1}. Our model consists of three parts: 1) {\bf Point embedding module}: We characterize local point cloud feature sets at different scales as embeddings of points. We add the Local Point Attention of our design to the representation to make the model more sensitive to edge bulges. This preserves more detail and prevents the point cloud from being over-smooth.2) {\bf Transformer module}: We propose Sparse Encoding in the transformer to perceive the structural relationships of the point cloud. Sparsity is added to the encoding as a reference for sensitivity to outliers. 3) {\bf Output header module}: We design a tail output for the denoising task. Residual connections are added to facilitate the identification of noise. 

Note that the downsampling layer, which is often used in transformer models \cite{18,20,33}, is discarded from our network structure. This is due to the fact that we are working to recover surfaces that have been corrupted by noise. Moreover, our network has no requirement for the number of input points thus it can be applied to a wide range of scenes. We will show more details in the following sections.

\subsection{Point embedding} \label{Point embedding}
The purpose of this section is to obtain input for the Transformer module. For the transformer to work effectively on point clouds, the first step is converting the point cloud into a vector sequence. One of the most straightforward solutions is to use coordinates as a set of vector inputs. However, this approach can lead to the loss of local features. In order to enable each point to probe the geometric features of the local area in which it is located, a point embedding module has been designed. This also compensates for the deficiencies of Transformer in local feature extraction \cite{19,33}.

\begin{figure}[h]
	\centering{\includegraphics[scale=0.26]{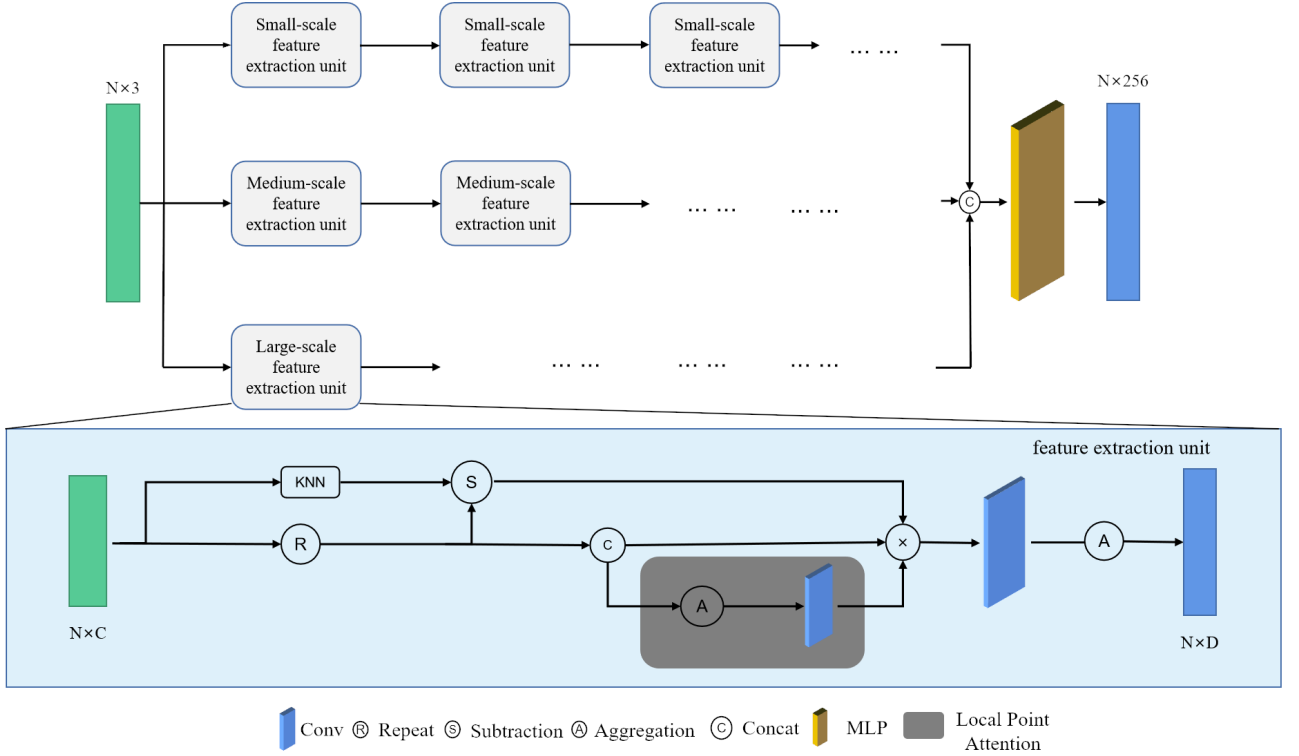}}
	\caption{Illustration of the point embedding module.}
	\label{fig2}
	\vskip-5pt
\end{figure}

In 2D images, convolution kernels of various sizes are usually designed to obtain the corresponding perceptual fields to extract features at different scales. Inspired by this experience, we choose to use the number of neighbor points to emulate the convolution kernels. As shown in Figure~\ref{fig2}, we design three feature extraction units with different numbers of neighbors in a parallel manner to capture features at various scales.

In the feature extraction unit, we construct K nearest neighbor points as a patch for each point in Euclidean space, followed by a convolution layer to generate high-dimensional features. The central point aggregates the patch as its own local features. After the several layers of feature aggregation, the output of the three feature extraction units is concatenated to obtain a vector sequence containing the local features of the point cloud. We call the above process {\itshape Point Embedding} and the resulting vector can be fed directly to the transformer.

 Formally, given the feature	${\bf F}^{l}_{p}=\{{\bf f}_{i}^{l}\}_{i=1}^{N} \in \mathbb{R}^{N \times \mathrm{d}^{l}}$ in the $lth$ layer, where $N$ denotes the number of points. Then the features of the $(l+1)th$ layer can be expressed as:
\begin{equation}
	\label{equa3}
	{\bf f}_{i}^{l+1}=G_{l}\left({\bf F}^{l}_{p}\right)=\max _{j \in N(i)}\left({\bf g}^{l}_{\theta}\left({\bf f}_{i}^{l}, {\bf f}_{j}^{l}-{\bf f}_{i}^{l}\right)\right)
\end{equation}
where ${\bf g}_{\theta}(.)$ represents a non-linear function with $\theta$ as a learnable parameter. $N(i)$ denotes the neighbors of point $i$, and $\max\left(.\right)$ is an aggregation function. 

To obtain local features at different scales, we set varying values of k in the three feature extraction units. The output is then stitched together, following a weight-shared multi-layer perceptron layer, which can be defined as:
\begin{equation}
	{\bf F}_{p.emb}=h_{\varphi}\left(\left[\left[{\bf F}^{1}_{p1}, \cdot \cdot \cdot {\bf F}^{4}_{p1} \right], \cdot \cdot \cdot \left[{\bf F}^{1}_{p3}, \cdot \cdot \cdot {\bf F}^{4}_{p3} \right]\right]\right)
\end{equation}
where ${\bf F}^{i}_{p1}$, ${\bf F}^{i}_{p2}$ and ${\bf F}^{i}_{p3}$ represent the outputs of the $ith$ feature extraction unit, and $h_{\varphi}(.)$ represents a weight shared multi-layer perceptron with $\varphi$ as a learnable parameter.$\left[. . .\right]$ represents the concatenation operation.

 \begin{figure}[h]
	\centering{\includegraphics[scale=0.25]{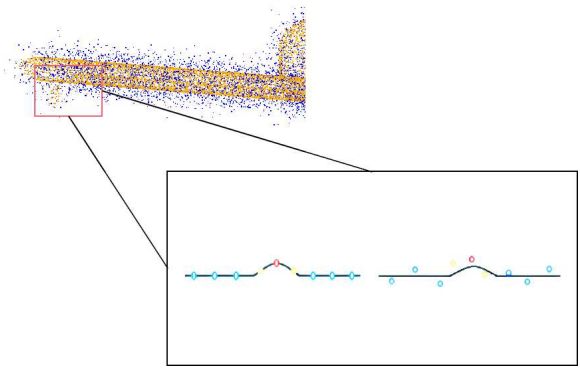}}
	\caption{Actual figure (top left) and schematic figure (bottom right). The orange points in the top left figure indicate clean points and the dark blue points indicate noisy points. As can be seen, it is almost impossible to identify the bulges in the noise points. In the lower right figure, the red point is easily recognized as noise causing the bulge edge to disappear. We increase the weights of the yellow points so as to retain the bulged detail.}
	\label{fig3}
	\vskip-5pt
\end{figure}

However, there are still problems with the above methods. A key issue in the denoising task is the balance between denoising and preserving protruding edges. This problem is unspecialised in most deep learning methods, and can lead to the network recognising the protruding edges as noise. This is especially so after several iterations, resulting in a loss of detail. Figure~\ref{fig3} illustrates this problem briefly.

To alleviate this situation, we introduced the Local Point Attention in feature extraction. This allows points that are on the same bump to be given more weight, thus preventing the object surface from being over-smooth.

Specifically, we compare the features of neighbor points at different scales to get the structural features where they are located. After concatenation, they are multiplied by a learnable matrix to obtain the weights. Our formula can be expressed as follows:
\begin{equation}		
	{\bf a}_{i}^{l+1}=A_{l}\left({\bf F}^{l}_{p}\right)=sigmoid\left({\bf W}^{l}_{a} \cdot \left(Agg\left({\bf f}^{l}_{i}, {\bf f}^{l}_{j}\right)\right)\right)
\end{equation}
where $Agg(.)$ represents the aggregation function. Then equation~(\ref{equa3}) can be rewritten as: 
\begin{equation}
	\label{equa4}
	{\bf f}_{i}^{l+1}=G_{l}\left({\bf F}^{l}_{p}\right)=\max _{j \in N(i)}\left({\bf g}^{l}_{\theta}\left(\left({\bf f}_{i}^{l}, {\bf f}_{j}^{l}-{\bf f}_{i}^{l}\right) \odot {\bf a}^{l}_{i}\right)\right)
\end{equation}
where ${\bf a}_{i}^{l}$ denotes the $ith$ point at $lth$ layer with the surface similarity weights of its neighbors compared to itself. We will show the specific effects in the ablation experiments.

\subsection{Sparse encoding} \label{spare encoding}
An advantage of the transformer is that each token can attention to any location information \cite{40}. Our model is built on the standard transformer and implements the point cloud denoising task. But the transformer \cite{27} was originally designed for sequence modeling and its self-attention mechanism only takes into account the semantic similarities between individual points. To make it better at capturing the structural relationships between points, we propose {\itshape Sparse Encoding}.

When describing the structure of a point cloud, the real directional orientation between points is only available in non-Euclidean space. This is difficult to compute for point clouds in Euclidean space. On the basis of ensuring the invariance of the overall model permutation, we use learnable coordinate differences as an approximation to the directions in non-Euclidean space. It is worth noting that we only make direction predictions for neighbor points. This is to allow the centroids to identify the spatial structure as well as to prevent errors that occur at longer distances.

For point cloud noise, the sparsity of the points is an important reference for judging the noise size and location information. We discard counting points as a criterion for evaluating sparsity with a more dynamic approach being chosen instead. We use the learnable inverse of the neighbor distance to evaluate the sparsity of points in range. This method not only allows the number of neighbors to be represented but also to some extent resembles the `degree' in a graph.

Specifically, given a coordinate vector set of points ${\bf X}=\left\{{\bf x}_{i}\right\}_{i=1}^{N} \in \mathbb{R}^{N \times d}$,where $N$ represents the number of points in the set and d represents the dimensionality of the vector set, generally we take $d=3$.The location code we have designed can be expressed as follows:
\begin{equation}
	{\bf encoding}_{i}=M_{\nu}\left( Concat \left({\bf x}_{i}-{\bf x}_{j},\left(\left\|{\bf x}_{i}-{\bf x}_{j}\right\|_{2}^{2}\right)^{-1}\right)\right)
\end{equation}
where $M_{\nu}$ represents a weight shared multi-layer perceptron with $\nu$ as a learnable parameter and $j$ represents neighbor points for $i$. In our experiments, we set the number of neighbor points to $3$.

Our point embedding and sparse coding ultimately form the input to our transformer. The purpose of point cloud denoising is to recover the noisy points to the underlying surface of the object, a task that transfers one set to another with the same properties. At the same time, in order to reduce the number of parameters and the calculation time, we, therefore, use only the encoder part of the transformer, simplifying the decoder part. Our module has a total of six transformer layers and each layer is a Pre-LN structure \cite{41}. Aimed at the organization of the point cloud, we redesign the Feed Forward Network. We use a weight-shared multi-layer perceptron to further extract features from the point cloud.
\begin{equation}
	\begin{aligned}
		{\bf F}^{l+1}_{o}=MLP&\big(LN\big( MSA \left(LN\left({\bf F}^{l}_{o}\right)\right)+{\bf F}^{l}_{o}\big)\big)+{\bf F}^{l}_{o}\\
		&{\bf F}^{0}_{o}={\bf F}_{p.emb}+{\bf encoding}
	\end{aligned}
\end{equation}
where $F^{l}_{o}$ represents the output of layer $l$, and $LN$ represents Layer Normalization. The multi-head attention mechanism plays a particularly crucial role in our task, describing the relationships between the points in different dimensions. Note that the order of inputs do not change in the above process. Meanwhile, the output has the same dimension and structure as the input, providing a great help to our loss function and the completeness of the point cloud. Since the calculations involved in the process are permutation invariant, a different order of inputs does not change the final output. As mentioned above, this is naturally suited to unordered point clouds.

\subsection{Output header} \label{output}
We specifically design output header module to handle the point cloud denoising task. The module consists of a densely connected multi-layer perceptron containing four linear layers. Each linear layer has a GELU nonlinearity. The module uses residual connections to transform the task into simulated noise for better denoising. The final output of our network is the denoised coordinates of each point. The computational process can be expressed as follows:
\begin{equation}
	{\bf Y}=P_{\epsilon}({\bf F}_{o})+{\bf X}
\end{equation}
where $X$ is the input coordinate set of the point cloud with noise and ${\bf F}_{o}$ represents the output of the transformer module. The output $Y$ represents the denoised coordinate.
\subsection{Loss function} \label{loss}
Choosing an appropriate loss function is particularly critical for the denoising task as it directly affects the denoising performance. We consider the differences between synthetic datasets and raw point clouds to design a two-part loss function in supervised training.

For quantifying the distance between the denoised point cloud and the ground truth point cloud, we adopt the Chamfer Distance(CD) \cite{42} as our loss function. This has also been shown to be effective by previous work \cite{12,14,15}. We make some changes to the CD to make it easier to visualize during the training process. The exact expression is as follows:
\begin{equation}
	\operatorname{Loss}_{CD}=\sum_{{\bf y} \in {\bf Y}} \sum_{{\bf x} \in {\bf X}} \min _{\substack{{\bf y} \in {\bf Y}\\{\bf x} \in {\bf X}}} \|{\bf x}-{\bf y}\|_{2}^{2}
\end{equation}
Nevertheless, it has been shown in previous work \cite{12,14} that the use of a relative minimum distance loss function leads to inferior visual effects. And this phenomenon was confirmed again in our experiments. Particularly at several iterations, the point cloud can appear filamentous or clustered. To enable points to move away from each other, we choose an `absolute' distance function as an additional loss function. Note that thanks to the point-to-point mapping nature of our network and the uniform distribution of points in the training set, we are able to choose this approach. Our specific loss function is as follows:
\begin{equation}
	\operatorname{Loss}_{AD}=\sum_{i=1}^{N}\left\|{\bf x}_{i}-{\bf y}_{i}\right\|_{2}^{2}
\end{equation}
where ${\bf x}$ represents the input coordinates and ${\bf y}$ represents the output coordinates. Overall, our loss function is as follows:
\begin{equation}
	\operatorname{Loss}=\alpha \cdot \operatorname{Loss}_{CD}+\beta \cdot \operatorname{Loss}_{AD}
\end{equation}
The above processes are all evaluated on uniformly distributed point clouds. However, raw point clouds are generally non-uniform. Given this case, we reduce the weight of the `absolute' distance function. We set $\alpha=0.9$ and $\beta=0.1$ in our experiments. By minimizing the overall loss function, we can ensure that the denoised points are restored to the underlying surface as closely as possible.

\section{Experiments} \label{sec4}
In this section, we evaluate our method on various datasets and compare it with other state-of-the-art methods quantitatively and qualitatively.
\subsection{Setup} \label{setup}
\subsubsection{Dataset} \label{dataset}
We use the training set from the paper \cite{15} and extract 100 different meshes in the ModelNet40 \cite{43} dataset for training. We sampled $10k-20k$ points at each mesh by Poisson sampling. Similar to previous work \cite{15,16}, we normalize the point cloud to the unit sphere, perturbed by Gaussian noise with standard deviations from $1\%$ to $4\%$ of the radius of the bounding sphere, and then split it into patches having $1k$ points as input.

For testing, we select $20$ meshes on each of the ModelNet40 \cite{43} and PU-Net \cite{44} datasets. we sample $10k$ points for each mesh and perturb them with Gaussian noise with standard deviations of $1\%$, $2\%$, and $3\%$ on the diagonal of the bounding box, respectively. To validate the effectiveness of the network structure we designed and because of the greater object details in the ModelNet40 dataset, we carry out ablation experiments on the ModelNet40 dataset and quantitative analysis on the PU-Net dataset. Furthermore, for validating the results of our model in different noise and raw point clouds, we add different types of noise to the data and use the Paris-rue-Madame \cite{45} dataset as the raw point cloud validation metric.
\subsubsection{Metric} \label{metric}
To quantitatively analyze the effect of denoising, we adopt two evaluation methods often used in previous work: the Chamfer Distance (CD) \cite{42} and the Point-to-Mesh Distance (P2M) \cite{46}. Where the Chamfer Distance describes the sum of the shortest distances of points between two point clouds:
\begin{equation}
	\operatorname{CD}=\frac{1}{\left|{\bf S}_{1}\right|} \sum_{{\bf x} \in {\bf S}_{1}} \min _{{\bf y} \in {\bf S}_{2}}\|{\bf x}-{\bf y}\|_{2}^{2}+\frac{1}{\left|{\bf S}_{2}\right|} \sum_{{\bf y} \in {\bf S}_{2}} \min _{{\bf x} \in {\bf S}_{1}}\|{\bf y}-{\bf x}\|_{2}^{2}
\end{equation}
where ${\bf S}_1$ and ${\bf S}_2$ represent two different point clouds, ${\bf x}$ and ${\bf y}$ represent the coordinates of the points.The Point-to-Mesh Distance, on the other hand, evaluates the denoising effect by calculating the average sum of the distances between the denoised point cloud and the clean mesh triangles.
\subsubsection{Implementation details} \label{id}
We implemented our model in pytorch \cite{47}. We used the Adam optimizer and set the learning rate of the Adam decay to 0.001 and the smoothing constants to 0.9 and 0.999 respectively. For 30K patches used as training input, we performed 200 epochs with an initial learning rate of 0.0005 and decreased to half of the original after 50 epochs. During testing, when the noise level was high, we iterated with reference to previous methods \cite{12,14,15} in order to obtain better denoising performance. We went through only one iteration for $1\%-2\%$ of the noise and two iterations for $3\%$.
\subsection{Comparison to state-of-the-art} \label{Ctsoat}
\subsubsection{Quantitative results} \label{quanr}
\begin{table}[h]
	
	\centering
	\caption{Comparison of denoising algorithms. Each data in the table is evaluated on 20 point clouds of different shapes selected from the PU-Net dataset.\label{table1}}{
		
	 \renewcommand\arraystretch{1.3}
	 \setlength{\tabcolsep}{4mm}{
			\begin{tabular}{c|cccccc}
				\hline
				Noise level & \multicolumn{2}{c}{1\%} & \multicolumn{2}{c}{2\%} & \multicolumn{2}{c}{3\%} \\
				Metric      & CD         & P2M        & CD         & P2M        & CD          & P2M       \\
				\hline
				Noisy       & 3.544      & 1.233      & 7.620      & 4.188      & 12.838      & 8.655     \\
				MRPCA       & 3.016      & 1.022      & 3.764      & 1.126      & 5.103       & 2.033     \\
				GLR          & 2.945      & 1.029      & 3.695      & 1.289      & 4.872       & 2.109     \\
				\hline
				PCNet        & 3.462      & 1.105      & 7.352      & 3.572      & 12.851      & 8.542     \\
				DMR          & 4.425      & 1.626      & 4.968      & 2.022      & 5.885       & 2.708     \\
				Score        & 2.427      & 0.429      & 3.545      & 1.027      & 4.795       & 1.988     \\
				\hline
				\bf{Ours}        & \bf{2.288}      & \bf{0.360}      & \bf{3.251}      & \bf{0.843}      & \bf{4.070}       & \bf{1.470}    \\
				\hline
			\end{tabular}
	}}
	
\end{table}
In this section, we quantitatively compare our method with some classical non-deep learning based point cloud denoising methods, as well as state-of-the-art deep learning based methods such as the MRPCA \cite{7} algorithm based on sparse representation, the GLR \cite{10} algorithm based on graphs, the PointCleanNet \cite{12} algorithm, which are the pioneers of deep learning methods, DMR \cite{15} and Score \cite{16} algorithms. To more comprehensively validate the effectiveness and robustness of our method, we add Gaussian noise, uniform noise, and Laplace noise to the test set, and then feed the test set directly into the algorithm to calculate the CD and P2M results between the output point cloud and the real point cloud, as shown in Table~\ref{table1}, in the range of $1\%$ to $3\%$ standard deviation of Gaussian noise, our method is significantly better than previous deep learning and non-deep learning methods. Moreover, our method outperforms previous methods in the case of large noise standard deviations. 

\begin{table}[h]
	\renewcommand{\arraystretch}{1.3}
	
	\caption{Comparison of denoising algorithms in different noises. We sampled 10K points from each point cloud.\label{table2}}{
		\setlength{\tabcolsep}{8mm}{
			\begin{tabular}{ccccc}
				\hline
				Noise        & \multicolumn{2}{c}{Laplacian noise} & \multicolumn{2}{c}{Uniform noise} \\
				Metrics      & CD               & P2M              & CD              & P2M             \\
				\hline
				Noisy        & 4.070            & 2.301            & 5.420           & 3.409           \\
				PCNet    & 3.899            & 1.986            & 5.134           & 3.124           \\
				DMR     & 4.013            & 2.208            & 5.242           & 3.281           \\
				Score  & 2.275            & 0.812            & 4.200           & 2.457           \\
				\hline
				\bf{Ours}         & \bf{2.126}            & \bf{0.752}            & \bf{3.801}           & \bf{2.100}          \\
				\hline
			\end{tabular}
	}}{}
	
\end{table}

Although the Gaussian noise is used in the training, Table~\ref{table2} shows that with uniform noise and Laplace noise, our method also has a significant advantage over other methods.

\subsubsection{Qualitative results} \label{qualr}
\begin{figure}[h]
	\centering{\includegraphics[scale=0.35]{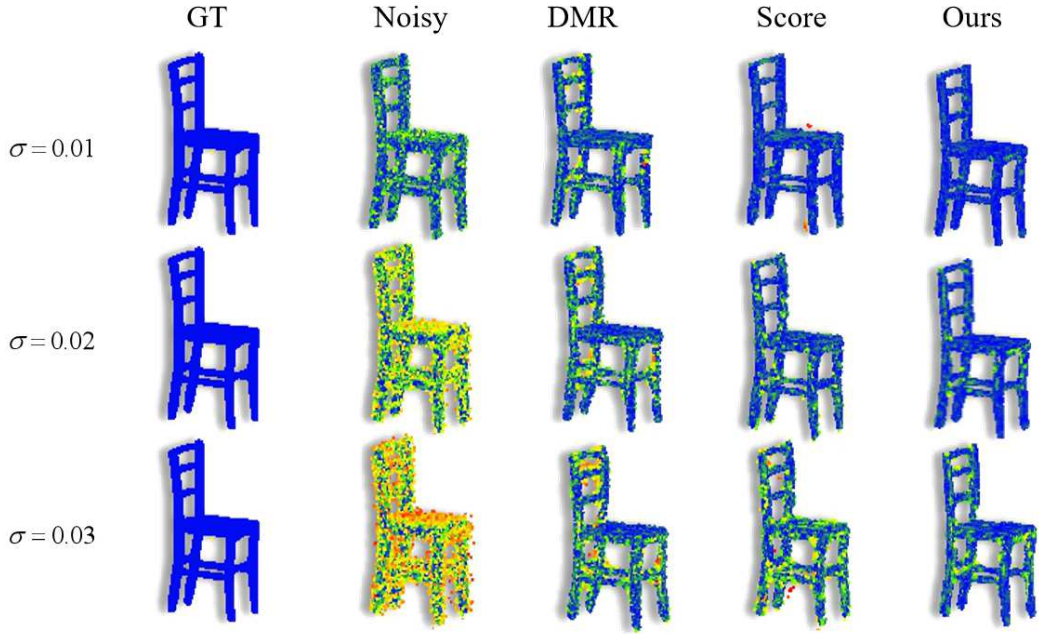}}
	\caption{Visual comparison of denoising methods. Blue points are on the surface of the object, the brighter the color, the further points are from the surface.}
	\label{fig4}
	\vskip-5pt
\end{figure}
\begin{figure}[h]
	\centering{\includegraphics[scale=0.3]{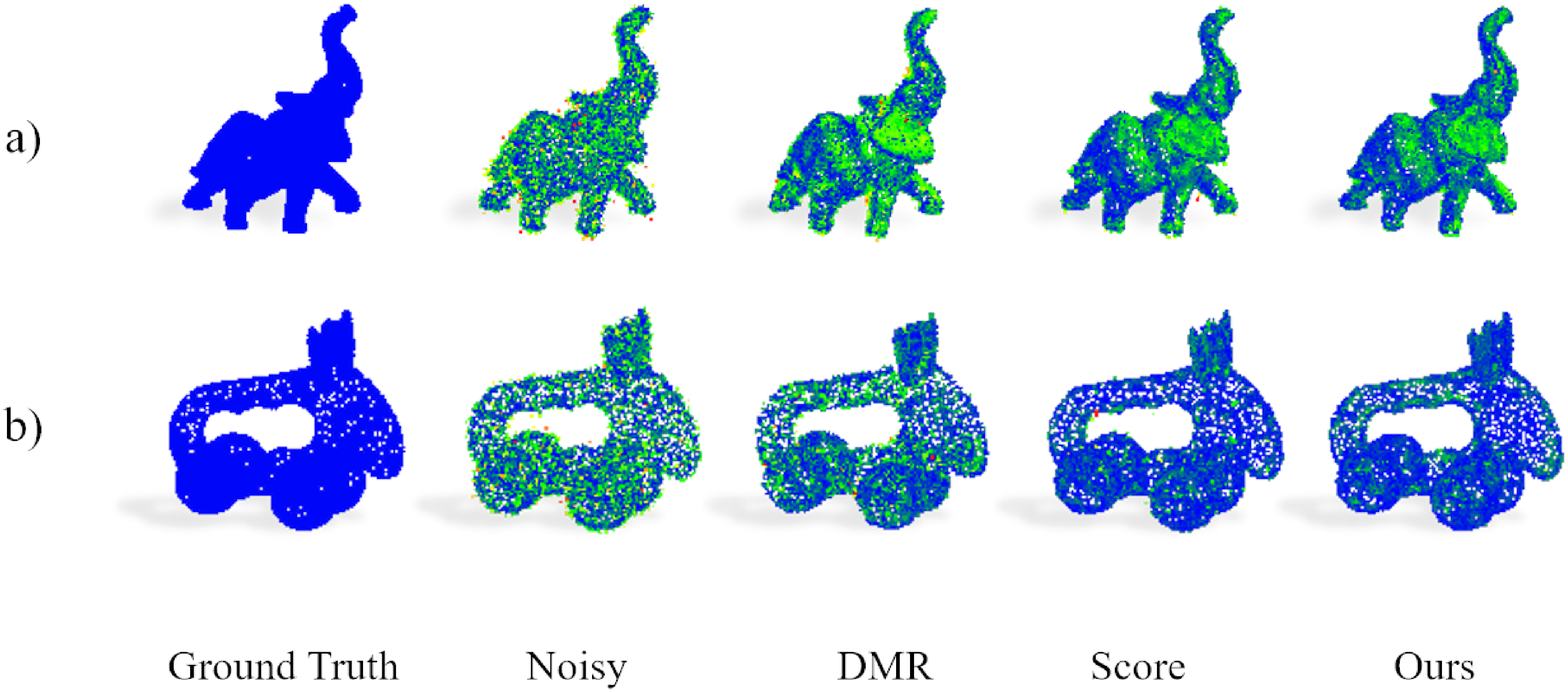}}
	\caption{Visual comparison of denoising methods with different noises. Where (a) denotes uniform noise and (b) denotes Laplacian noise.}
	\label{fig5}
	\vskip-5pt
\end{figure}

We show in Figure~\ref{fig4} the distribution of points on the surface of the object after denoising by the current state-of-the-art DMR and Score methods compared to our method for different levels of Gaussian noise interference. We have colored all the points according to distance from a clean object surface, where blue indicates proximity to the surface and brighter means further from the surface. It is clear to observe that our method outperforms DMR, the output points are closer to the object surface and, unlike Score, our method does not produce outliers. In particular, it has a more pronounced visual effect when the noise level is low; it is also able to bring the noise points as close to the object's surface as possible when the noise level is particularly high. Meanwhile, we have evaluated several methods in different noises and, as shown in Figure~\ref{fig5}, our method works well even in noises that are not identified by the model.

\begin{figure}[h]
	\centering{\includegraphics[scale=0.21]{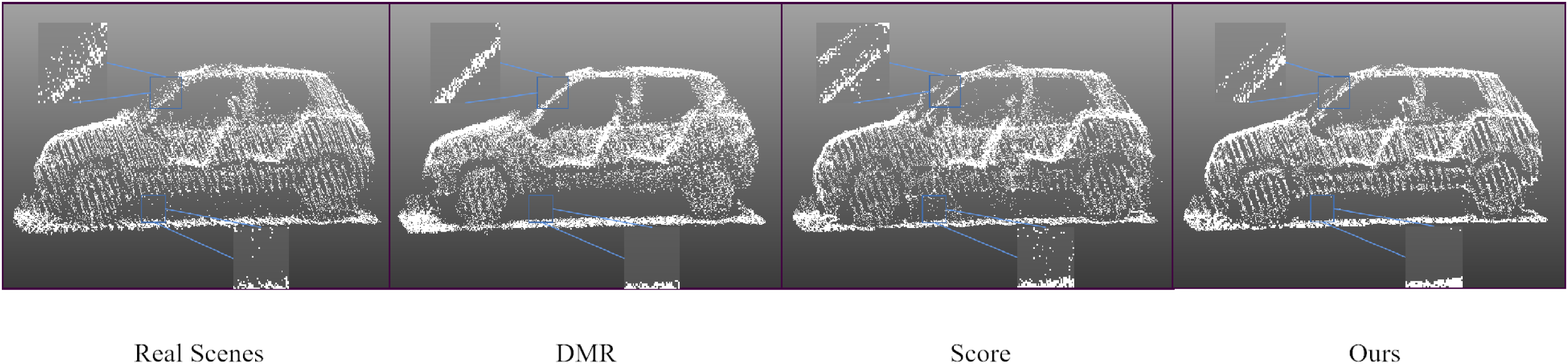}}
	\caption{Visualisation of raw point cloud denoising. Our method performs well in terms of noise removal while preserving the integrity of the vehicle's A-pillar.}
	\label{fig6}
	\vskip-5pt
\end{figure}

We have further compared the above methods in the raw point cloud. We have used the Paris-rue-Madame \cite{45} dataset as a reference. As the clean point cloud is unknown, it is not possible to quantify the deviation from noise and we can only analyze the dataset qualitatively. The Paris-rue-Madame dataset consists of two files, each with 10 million points. We only take the coordinates of each point, and given the limited computing power of our GPU, we divide the 10 million points into 1000 parts, convert them to the size of a unit sphere and feed them into the algorithm, and then perform the stitching operation after obtaining the output. For qualitative analysis, we take a portion of the dataset and show the results after one round of iterations in Figure~\ref{fig6}. It can be seen that our method retains more detail than the DMR, and the denoising effect is better than the Score method, in that the denoised points are closer to the surface of the object, and the denoised surface is smoother.

\subsubsection{Ablation studies} \label{as}
\begin{table}[h]
	\renewcommand{\arraystretch}{1.2}
	\caption{Comparison of ablation studies. In the first experiment, we replaced the self-attentive layer with a convolution layer. In the third experiment, we use the learnable point coordinates as the location embedding.\label{table3}}{
		
		\setlength{\tabcolsep}{10mm}{
			\begin{tabular}{cccc}
				\hline
				Component          & 1\%   & 2\%   & 3\%   \\
				\hline
				No self-attention  & 2.501 & 3.567 & 4.452 \\
				No pos.encoding    & 2.416 & 3.350 & 4.119 \\
				Coor.encoding      & 2.315 & 3.328 & 4.088 \\
				Attention+Spare & 2.288 & 3.251 & 4.070 \\
				\hline
			\end{tabular}
	}}{}
\end{table}
\begin{figure}[h]
	\centering{\includegraphics[scale=0.07]{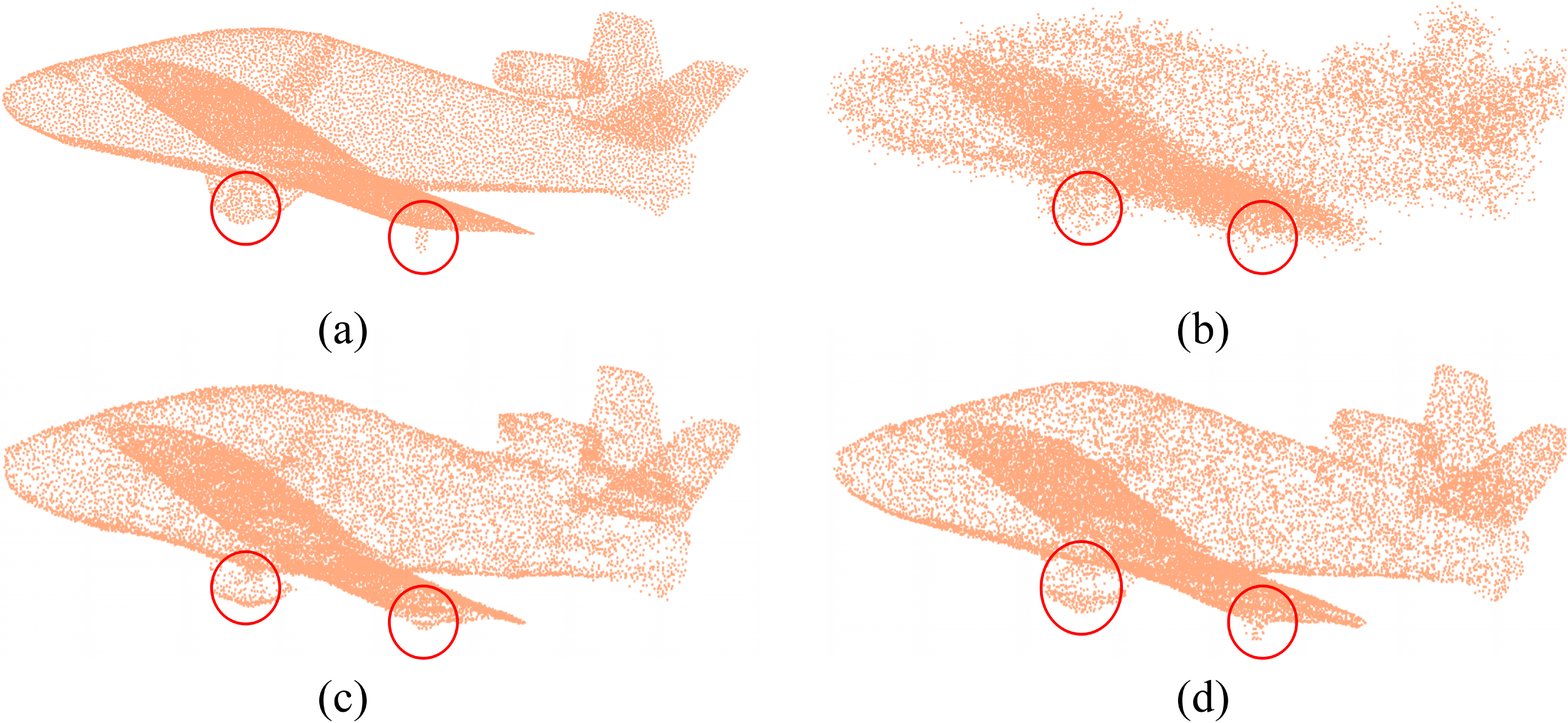}}
	\caption{Visual Ablation Experiment of Local Point Attention. (a): clean (b): noisy (c): no local point attention (d): ours. It can be observed that after three iterations, the flaps of aircraft (c) have significantly disappeared, while aircraft (d) is still present.}
	\label{fig7}
	\vskip-5pt
\end{figure}

We focus in this section on showing the reasonableness of our network structure design, as shown in Table~\ref{table3} which demonstrates the impact of our design network part structure on the final denoising effect. The first experiment (first row) discards the self-attentive layer and uses a convolution layer instead, to verify the effect of the self-attention mechanism on our experiments. The second and third experiments (second and third rows) change the positional encoding in the transformer module (the second experiment has no positional encoding) to verify the effectiveness of our position encoding design. We used a point cloud of 10K points with $1\%-3\%$ Gaussian noise in our experiments.

Each module of our design contributes to denoising performance. It is worth noting that the role of self-attention increases as the noise gradually increases. The position encoding plays a larger role in the less noisy cases. We believe that it is when the noise is high that the offset of the position can lead to an uncontrolled increase in the distance between points, affecting the effect of positional encoding. This problem can be mitigated by using different scales of position encoding.

To verify the effectiveness of validating our local attention, we conducted experiments on the ModelNet40 \cite{43} dataset. Figure~\ref{fig7} qualitatively shows the results of our experiments. At $2\%$ noise level, after $3$ iterations, the aircraft wing tail bulge is smaller or even disappears. However, this is significantly improved after increasing the Local Point Attention.
\section{Conclusion} \label{sec5}
In this paper, we propose a new denoising architecture, NoiseTrans. We formulate the point cloud as a set of vector sets and successfully merge the point cloud denoising task with transformer through a number of technical innovations. Different from previous work, our model extracts local features at diverse scales and captures the semantic relationships between points and structural features with the help of the transformer encoder. To preserve details in denoising, we introduce changes to the embedding vector. It can be shown from a number of experiments that our model achieves state-of-the-art performance on different datasets. 

At present, there is still no standard dataset for the point cloud denoising task, which also causes evaluation discrepancies. We hope to witness the availability of large standard datasets for denoising in the future, allowing further refinement of our approach.

\bibliography{reference}

\end{document}